# On the Importance of Sign Labeling: The Hamburg Sign Language Notation System Case Study

**Maria Ferlin**
Gdańsk University of Technology
Gdańsk, Poland
maria.ferlin@pg.edu.pl

**Sylwia Majchrowska**
AI Sweden
Gothenburg, Sweden
sylwia.majchrowska@ai.se

**Marta Plantykow**
Women in AI
Gdańsk, Poland
m.plantykow@gmail.com

**Alicja Kwasniewska**
SiMa Technologies
San Jose, CA, USA
alicja@sima.ai

**Agnieszka Mikołajczyk-Bareła**
VoiceLab, NLP Lab
Gdańsk, Poland
agnieszka.mikolajczyk@voicelab.ai

**Milena Olech**
Intel Corporation
Gdańsk, Poland
milena.w.olech@gmail.com

**Jakub Nalepa**
Silesian University of Technology
Gliwice, Poland
jnalepa@ieee.org

## Abstract

Labeling is the cornerstone of supervised machine learning, which has been exploited in a plethora of various applications, with sign language recognition being one of them. However, such algorithms must be fed with a huge amount of consistently labeled data during the training process to elaborate a well-generalizing model. In addition, there is a great need for an automated solution that works with any nationally diversified sign language. Although there are language-agnostic transcription systems, such as the Hamburg Sign Language Notation System (HamNoSys) that describe the signer's initial position and body movement instead of the glosses' meanings, there are still issues with providing accurate and reliable labels for every real-world use case. In this context, the industry relies heavily on manual attribution and labeling of the available video data. In this work, we tackle this issue and thoroughly analyze the HamNoSys labels provided by various maintainers of open sign language corpora in five sign languages, in order to examine the challenges encountered in labeling video data. We also investigate the consistency and objectivity of HamNoSys-based labels for the purpose of training machine learning models. Our findings provide valuable insights into the limitations of the current labeling methods and pave the way for future research on developing more accurate and efficient solutions for sign language recognition.

## 1 Introduction

Deep learning (DL) algorithms, benefiting from automated representation learning, are widely employed to extract knowledge from vast amounts of data in a wide range of applications, including healthcare [1], waste management [2, 3], and natural language processing [4, 5, 6]. Powerful graphics processing units, tensor processing units, and high-quality videos, images, and texts are essential for DL-based software to perform at a human level across a variety of tasks [7]. Whereas, machine learning (ML) algorithms utilize manually hand-crafted feature extractors commonly followed by feature selectors to pick the most informative features, therefore, it is worth noticing that even the best ML algorithms struggle to train properly in the absence of high-quality, well-annotated data. Data-centric (DL) techniques are becoming more widely used in different real-world applications, and sign language recognition [8, 9] is not an exception here. Novel technologies may help with translation between sign languages and verbal languages and thus facilitate interpersonal communication. Unfortunately, many publicly available sign language corpora are small, biased, and sometimes impure in terms of labels associated with a data entry being incorrect or only partially correct [10, 11].

A major challenge regarding these translations is the lack of one, universal sign language (SL) – they vary widely from country to country and even from region to region within a single country [12, 13, 14, 15, 9, 16, 17]. In verbal languages, there is a written representation of each word, which consists of conventional single characters – letters – that can differ according to national alphabets. In the case of sign language translation, each sign (gesture) must gain written representation, which traditionally is a word of specific spoken language with a similar meaning to the sign meaning. They are called glosses [18]. In the last 40 years, several unified alternatives, based on specific characteristics of sign movements in space, appeared in the international community. For instance: Stokoe notation [19], SignWriting [20], and Hamburg Sign Language Notation System (HamNoSys) [12, 21], further converted into an XML form known as Signing Gesture Mark-up Language (SiGML) [22]. However, until now, none of these options are well known to the Deaf community, and therefore no uniform procedure for automatic labeling multilingual data has been developed [10, 11, 23].

Our research led us to the question whether HamNoSys-based labels are sufficient for training machine learning models. Utilizing the open-source sign language corpora annotated by linguists in the universal phonetic HamNoSys, we explored possible challenges that HamNoSys users may encounter in the process of labeling video data. To address this research gap, we thoroughly evaluated the provided labels and annotations guidelines in terms of their consistency and objectivity.

We managed to identify the signer's initial position frame for detailed analysis of the signer configuration in contrast with the HamNoSys notation. Next, we investigated the distance between body parts for various HamNoSys symbols. We also carefully analyzed the non-numerical features such as choice of dominant hand and subjective bias. Additionally, we investigated the impact of the HamNoSys symbols reduction on reversal translation.

Overall, the contribution of this paper is twofold. Firstly, we presented an overview of the properties of existing open sign language corpora using a self-developed parser for HamNoSys labels. Secondly, we identified key challenges and inconsistencies in the HamNoSys annotation guidelines that can lead to machine learning translation system malfunction.

The remainder of the paper is organized as follows. In Section 2, we contextualized our research within the current state of the art in the field. Subsequently, in Section 3, we presented the materials and methods employed in the paper, as well as an analysis of the existing open sign language databases. The challenges and observations identified while working with HamNoSys are gathered in Section 4. While in Section 5 we conducted a discussion about problems and possible solutions for HamNoSys-based labels for machine learning purposes. Finally, in Section 6 we drew conclusions and outlined the future work in this area.

## 2 Background

Technological advances over the past two decades have enabled the barriers encountered by deaf people to diminish. The way people can communicate and learn has significantly improved with innovations introduced to both hardware and software. Due to inconveniences caused by the need of using specialized equipment, e.g., gloves [24], the preference is to lean towards camera-based solutions. They impose neither any specific action on the user nor a need to wear any additional sensors that may affect daily activities, as in the case of, for instance, wearable inertial or electromyography-based hand movement detection solutions [25].

Computer Vision based solutions for hand movement tracking have been well-studied for over 30 years. Starner *et al.* [26], initially proposed a color-based system for hand movement tracking, and later improved it to remove the need for glove-wearing. Later, another system was introduced by Glenn *et al.* [27]. Although it allowed for conversion to both written and spoken English, it suffered from two major disadvantages. First of all, it was only targeting American Sign Language (ASL) and secondly, it worked in a controlled environment. Simultaneously, studies on improvement in detection accuracy were conducted, such as multi-flash camera casting shadows along depth continuities in the hand shape [28].

At the same time, machine learning methods have been actively studied and applied to the sign recognition task. Some initial works were based on neural networks [29] or Hidden Markov Models [30]. A wider review of vision-based gesture recognition was carried out by Wu and Huang [31]. The fact that these works were recognizing gestures, not vocabulary, and very often were limited to some pre-defined segmented words only limits the possibility of their usage in the SL recognition task. Recently, due to the rapid development of deep neural networks, a strong emphasis was put on such systems. Zheng *et al.* [32], proposed to – at first – pre-process the recorded sequence with a frame stream density compression (FSDC) that aimed at reducing redundant images, and then to process them using neural machine translation modules that consisted of temporal convolutions and Bidirectional Gated Recurrent Units (BiGRU). Another example included the use of Generative Adversarial Networks (GANs) for sign language gloss recognition based on spatial and temporal features extracted from video sequences [33]. The importance of contextual information in deaf-to-deaf and deaf-to-hearing communication was also analyzed in this work. GANs have been also used in the translation of signs into so-called sign pose motion graphs [34] resulting in promising accuracy scores.

However, most existing studies target only a specific national diversified finger-spelling system, e.g., ASL [35], Portuguese [36], Arabic [37], and others, and cannot be easily extended to other sign languages. Hamburg Sign Language Notation System (HamNoSys) was developed to alleviate this limitation [12, 13, 21]. Recent efforts to use HamNoSys concentrated on enlarging multilingual lexicons and improving the labeling process. The main focus was put on the annotation process [38, 39, 40], as the grammar of the notation system is complex and well-known only to sign language experts. Despite the fact that not all existing Sign Language (SL) corpora are annotated using HamNoSys, this unified annotation system has drawn the interest of numerous researchers [22, 41, 42, 43, 4] due to the amount of multilingual research it may facilitate.

More lately, SL-related research in HamNoSys applications focused on automatic translation HamNoSys into sign language by representing such input in a graphical form, e.g., based on avatars [44, 5]. Some of the solutions that apply this technology were implemented to use a predefined corpus that maps a written word into the HamNoSys notation, so that the word itself can be used as an input for the system [45, 46]. The most recent research using the HamNoSys notation aimed to take advantage of sign language resources available on social media and group segmented phonemes without access to the transcription [11].

There are only a few studies that attempted to use HamNoSys-based classification to translate a sign into text. Researchers were investigating the ability to use HamNoSys for SL-unified sub-unit classification [41, 47, 48], which can be a first step for SL recognition. In 2018, Grif *et al.* [49], proposed the work on Sign Language Gesture Recognition Framework Based on HamNoSys focused on five main features such as trajectory, orientation, axis, rotation, and handshape, using a tracking software library. On the other hand, the issue of HamNoSys annotations being in their current form insufficient for the DL-based model training was already issued in some works [23, 41, 6]. We followed this research pathway and comprehensively investigate this topic.

## 3 Materials and methods

In this section, we presented the materials and methods used in our research. We provided a brief overview of the Hamburg Sign Language Notation System (HamNoSys) (Section 3.1), the utilized sign language corpora (Section 3.2), the HamNoSys parser used in the annotation analysis (Section 3.3), and the body landmark model from Mediapipe employed to determine the coordinates of body parts (Section 3.4).

### 3.1 Hamburg Sign Language Notation System

The Hamburg Sign Language Notation System (HamNoSys), in linguistic terminology, is a phonetic transcription system that reflects posture and movement, making it language-independent [12]. In addition, the HamNoSys 4.0 from 2004 incorporates practical experience in using the system for languages other than American Sign Language (ASL), increasing its potential for international applications [13, 50]. We based our analysis on this particular version.

HamNoSys describes signs in terms of hand shape, hand position/orientation, and hand movement using a unique alphabet with an inventory of about 200 symbols, depending on its version. The notation can be divided into six basic building blocks. Typically, the signs written in HamNoSys notation consist of symbols in the order, presented in the upper panel of Figure 1 (the optional blocks are dashed), given that some blocks may not be present or may assign any intermediate position between those two (represented by appropriate symbols separated by \). The symbols combined in this way form the HamNoSys label for the given sign.

Four of the blocks, i.e., non-manual features, hand shape, hand position, and hand location describe the signer's initial configuration, as follows [51, 52]:

- Symmetry operator (optional) – denotes two-handed signs and determines how attributes should be mirrored and by which axis. The HamNoSys 4.0 defines 8 symmetry operators.
- Non-manual features (optional) – represent the features that can be used to describe the facial expression of a given sign (e.g., puffed or sucked-in cheeks); these are complementary diacritical symbols.
- Hand shape – refers to the hand shape description, which is composed of three sub-blocks – *base form, thumb position*, and *bending.* The HamNoSys 4.0 introduces 12 standard hand shapes (flat, fist, pointing index finger, etc.), which can be combined with diacritic signs for 4 thumb positions and for 6 bending types of individual fingers.
- Hand position – describes the hand orientation using two sub-blocks, namely the *extended finger direction*, which determines the 26 directions of the metacarpal of the middle finger, and the *palm orientation* with 8 possible positions defined relatively to the *extended finger direction.*
- Hand Location (optional) – consists of 47 symbols, which can be divided into three categories, the *location left/right* specifies the $x$ coordinate from the signer's perspective, the *location top/bottom* for the $y$ coordinate given by body location character, and the *distance* determines the $z$ coordinate in relation to the body part provided by top/bottom location component. If the sign in its initial position is located in a neutral space in front of the upper body part, the blocks can be omitted in the HamNoSys transcription.
- Movement – represents a combination of the movements on the path, which determines the target/absolute (location) or relative (direction and size) movements. This block is quite complex as the movement of the hand can be straight (26 directions), curved (8 different ways), circular, or directed to a specific location. Moreover, the movements can be combined sequentially or in parallel.

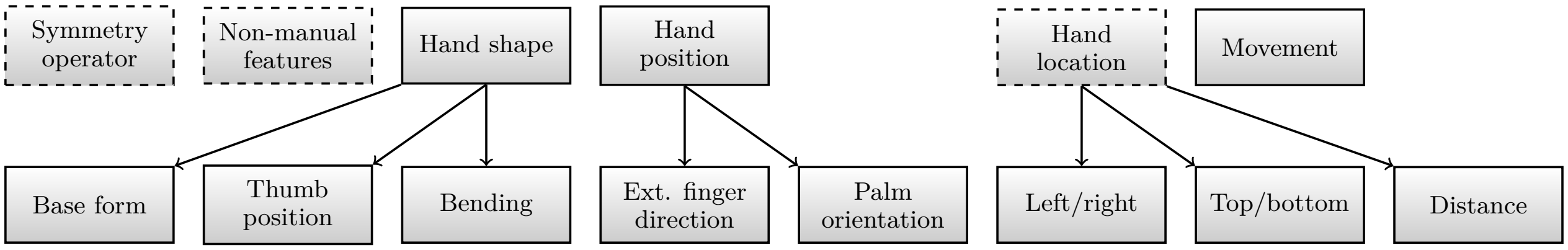

Figure 1: A detailed HamNoSys structure. The components in the boxes with the dashed border are optional, i.e., symmetry operator, non-manual features and hand location.

### 3.2 HamNoSys-annotated sign language corpora

As HamNoSys annotations are not language-dependent, they can facilitate cross-language research. Moreover, merging several multilingual corpora can effectively increase the number of available training data. Hence, a large amount of adequately labeled data can improve the generalization of machine learning systems over unseen data and prevent overfitting, i.e., memorizing a limited training set.

Combining different databases together also has its drawbacks and underlying challenges directly concerned with that. Firstly, many existing databases are collected over a period of years, which is of great importance in the case of recordings of sign language translators, as the appearance of the same person changes significantly. Secondly, the year the database was created can also indicate the HamNoSys version used to annotate it. When merging such datasets, some inconsistencies may appear due to notation variations. In addition, data from various sources were collected using different cameras, resolutions, or recording modes (from side, top or front view). Usually, they were coherent within one dataset but varied in general. Such diversification is desired in the case of large databases, but in the case of smaller ones may lead to inconsistencies and further poor performance of the system. The relevant drawback is the fact that datasets were recorded only indoors with considerable variation in artificial lighting. A system trained on such data could not be used outdoors with satisfying performance. In turn, the signers are commonly young and middle-aged adults. This might lead to problems with the proper recognition of signs performed by elderly people or with different origins. The signers were both in standing and sitting positions, which is a virtue regarding data diversification.

Nevertheless, all these factors may affect the machine learning model's ability to recognize a gesture, despite the quality of the labels themselves.

At the same time, most of the openly available data is offered in the form of online dictionaries – one example per gloss, which is a gesture of a signed word. That makes it extremely difficult to train the model in a fully supervised manner. Also in this form, the data must be scrapped from the website, which may be a source of errors (i.e., mismatched annotations). Moreover, they contain only isolated glosses without context. Dictionary signs are different from those used during the conversation. Such differences between dictionary and conversational notation block reduce the possibility of using them in everyday situations, even if the model would be sufficiently well-trained. A comprehensive overview of the existing corpora used in this analysis supplied with HamNoSys labels is provided below and summarized in Table 1.

**GLEX**: The Fachgebärden Lexicon – Gesundheit and Pflege [13] was developed at the German Sign Language Institute of the University of Hamburg between 2004 and 2007 and contains more than 2k signs, resulting in nearly 1.5h of recordings. The signers were depicted on a smooth, light blue background. The videos were presented in the same resolution of 108×80 pixels with a fixed fps of 25, and an average length of 2 seconds. This dataset was dedicated to technical terms related to health and nursing care. The synonymous terms and some common abbreviations, from which reference was made to the respective technical term, were provided in the online dictionary.

**GALEX**: The Fachgebärden Lexicon – Gärtnerei und Landschaftsbau [13] was also created at the Hamburg Institute of German Sign Language between 2006 and 2009. The videos were characterized by 160 pixels displayed across the screen horizontally and 90 pixels down the screen. The glosses were recorded with two translators – a woman and a man – against a plain blue background. This dataset consists of a few hundred signs (half an hour of recordings) related to technical terms for landscaping and gardening. The dictionary was intended to support deaf people in training for horticultural professions such as gardeners, landscapers, landscape architects, and horticultural engineers.

**BL**: The Basic Lexicon is part of the multilingual[1] three-year research project DICTA-SIGN [14], carried out since 2009 by a Consortium of European Universities[2]. The main purpose of creating the dataset was to show similarities between different sign languages, hence the same glosses were repeated in different languages. However, the individual sub-parts differ significantly in the duration (ranging from 2 to 100 seconds), resolution (reaching an average of 320×200 pixels), and style of recorded videos. For example, in the case of the French subpart, the videos are a combination of 3 views – front, side, and top – which necessitated early prepossessing involving cutting the frames into 3 parts. For the Greek part of the dataset, the director used the fade-in/out technique at the beginning and end of each video, respectively. About 1k signs were provided for each SL, resulting in half an hour of recordings for each. The shared list of glosses covered the topic of traveling.

**CDPSL**: The Corpus-based Dictionary of Polish Sign Language (PJM) [15] was created in 2016 at the Sign Linguistics Laboratory of the University of Warsaw. The dictionary was developed based on the PJM corpus, which collected the video data of 150 deaf signers who use PJM. About 11 hours of annotated recordings are available online. The dictionary interface can search for a sign by semantic properties such as handshape, localization, and additional features. The collected glosses were recorded in high resolution (1280×720 pixels) with two interpreters - a man and a woman - against a smooth gray background. In the upper right corner of the videos is the logo of the Polish Sign Linguistics Laboratory. The average video duration is 5 seconds. CDPSL gathers the most used PJM glosses.

**GSLL**: The Greek Sign Language Lemmas [9, 16, 17] was developed by the National Technical University of Athens and supported by the EU research project DICTA-SIGN. This is the only one of the databases listed here that is available for direct download without the need to scrap the videos one by one from the website. The collection was provided in form of selected separate frames. The declared number of frames per second equals 35. The dataset was dedicated to isolated SL recognition and contains a few hundred glosses signed by two participants, where each sign is repeated between 5 and 17 times. Finally, the glosses were recorded through a total of over 3k videos containing 161k frames.

[1] British Sign Language (BSL), German Sign Language (DGS), Greek Sign Language (GSL), and French Sign Language (LSF)

[2] Institute for Language and Speech Processing, Universität Hamburg, University of East Anglia, University of Surrey, Laboratoire d'informatique pour la mécanique et les sciences de l'ingénieur, Université Paul Sabatier, National Technical University of Athens, WebSourd

Table 1: Statistics for analyzed sign language datasets supplied with HamNoSys annotations.

| Dataset | Year | Languages | Topics | # glosses | # videos | # signers |
|---|---|---|---|---|---|---|
| GLEX | 2004 – 2007 | DGS | health | 723 | 829 | 2 |
| GALEX | 2006 – 2009 | DGS | landscaping | 514 | 568 | 2 |
| BL | 2009 – 2012 | BSL, DGS, GSL, LSF | traveling | 3078 | 4 123 | > 6 |
| CDPSL | 2012 – 2014 | PJM | everyday use | 2480 | 2 835 | 2 |
| GSLL | 2009 – 2021 | GSL | everyday use | 300 | 3 476 | 2 |

### 3.3 HamNoSys parser

As the information given by the HamNoSys label is primarily provided in form of a list of symbols, at first, it must be correctly encoded to make it more suitable for machine learning-powered classification. Here, we utilized self-developed HamNoSys parser [3] that assigned a HamNoSys character the corresponding numerical label [53]. The tool processed a given list of HamNoSys labels one by one by searching for the character describing the HamNoSys blocks mentioned in Section 3.1, and the block's classes were assigned in a step-by-step manner. If the symbol describing the mandatory building block was not found in the HamNoSys label in a proper position, then the whole label was considered incorrect and therefore omitted.

The parser did not process all of the HamNoSys characters – it discards two main blocks, namely *non-manual feature* and *movement*. The *non-manual feature* block added in the HamNoSys 4.0 was discarded in the parser as it was not fully developed and does not describe all possible features [54]. In addition, it is often not used by annotators (as it is optional). However, the non-manual features, particularly facial expressions, are reported to be the important features of a sign itself and should be considered. In turn, the entire block representing possible combinations of *movement* does not have a strictly defined structure. It is represented by arrows and additional symbols attached to characters representing other blocks. For this reason, it is challenging to automatically decode it. Moreover, the symbols – numbers – specifying particular fingers were also dropped. In many cases, it was not visible which specific fingers of a given hand were involved in the gesture due to the low resolution of the video. Therefore, this information was considered too detailed and disorganized and excluded from the analysis. Table 2 provides an overview of the HamNoSys symbols encoded into numerical, computer-friendly labels [54].

Table 2: The HamNoSys symbols recognized by the HamNoSys parser describing the signer's initial configuration. Characters describing the *thumb position*, *bending* and *hand location left/right* are shown together with additional base form symbol for better visualization.

| HamNoSys block | | Symbols | # classes |
|---|---|---|---|
| Symmetry operator | | [illegible] | 9 |
| Handshape | Base form | [illegible] | 12 |
| | Thumb position | [illegible] | 4 |
| | Bending | [illegible] | 6 |
| Hand position | Extended finger direction | [illegible] | 18 |
| | Palm orientation | [illegible] | 8 |
| Hand location | Left/right | [illegible] | 5 |
| | Top/bottom | [illegible] | 36 |
| | Distance | [illegible] | 7 |

### 3.4 Human pose estimation with MediaPipe Pose

The assessment of the consistency of the HamNosys notation is critical because it relies on a number of geometrical properties, such as locations and distances between certain body parts. In this way, the potential use of notation in automated systems trained to produce predictions in a supervised manner – that is, employing pairs of inputs and ground-truth annotations – can be evaluated. The model will not converge if the geometrical information overlaps many labels and cannot be easily distinguished. In order to investigate the usability of HamNoSys-based labels for machine learning purposes, the body landmark model from

[3]https://github.com/hearai/parse-hamnosys

Mediapipe [55] was used in the study. It is a neural network trained to determine the coordinates of body parts. Predictions produced by this network correspond to the locations of the 33 key landmarks.

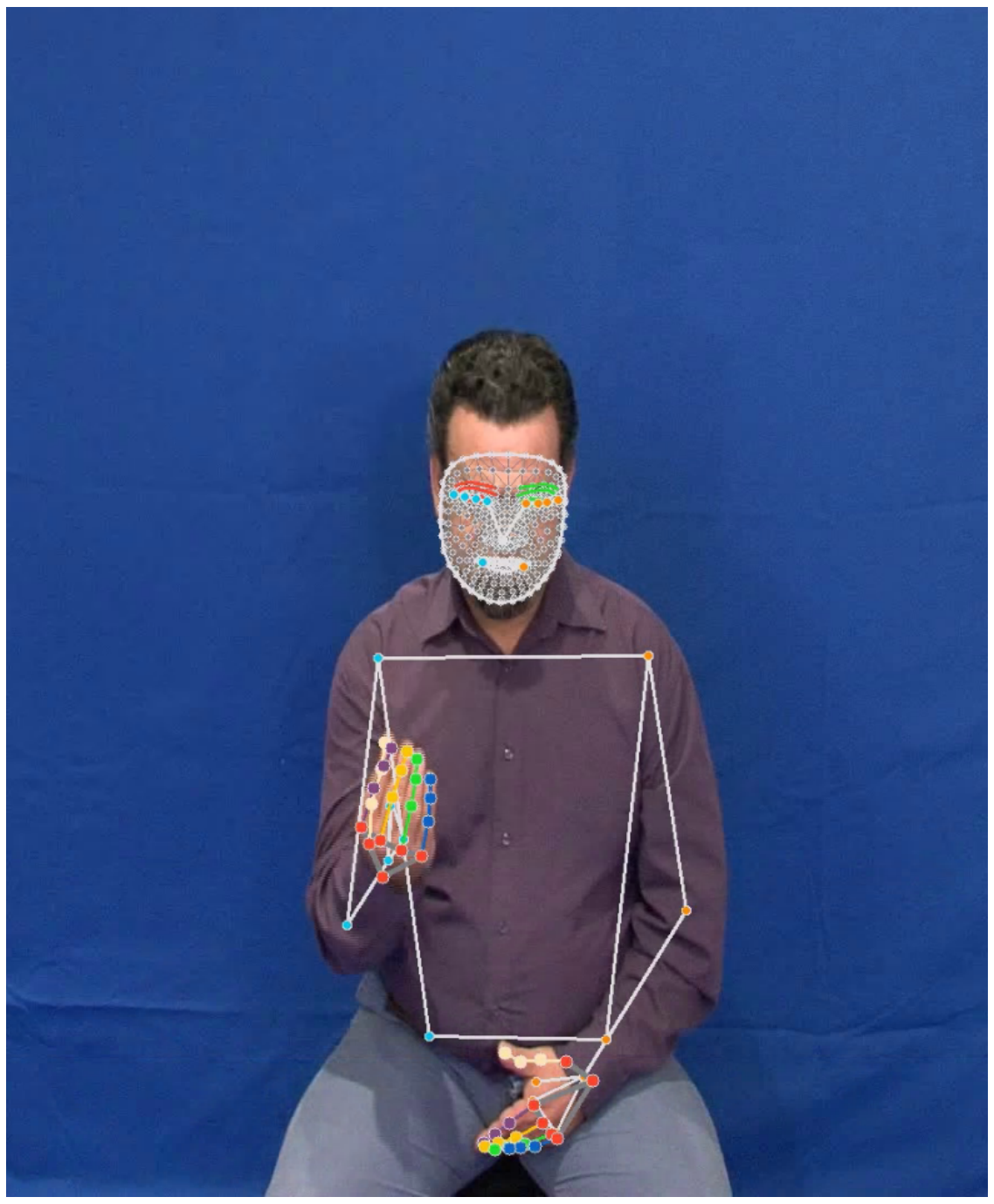

Figure 2: An example of the sign language gesture from GSLL dataset annotated with Mediapipe body landmarks. The colorful dots represent particular joints that are being followed, all are connected to create a pose scheme.

## 4 HamNoSys-based labels analysis

In this section, we investigated the challenges related to the labeling of the video using the HamNoSys annotation guidelines for sign language processing and translation purposes. We considered various aspects of the notation system that may reflect difficulties in automating this process.

### 4.1 Identification of the signer's initial configuration

Capturing non-manual features, the shape, the orientation, and the location of the hand must be related to the signer's initial position. Unfortunately, HamNoSys does not define any rules for the procedure of selecting the video frame indicating the initial configuration of the sign, implicitly assuming that the initial position is known to the annotator based on the knowledge of the analyzed sign language.

This creates ambiguity and introduces inconsistency for the computer-aided annotation process based on HamNoSys notation.

The first step in the annotation process is to analyze the video sequence and decide which video frame should be considered the one with the initial sign configuration. Since there are no defined guidelines, this decision depends on individual interpretation and may differ depending on corpora and annotators. One interpretation could be to consider the inflection point of the movement (if there is one). It is still possible to use different techniques to detect the inflection point and therefore determine slightly different initial frames. This, again, could lead to possible inconsistencies in annotations provided by different sources.

To better understand this problem, Fig. 3(a) shows the eight initial frames of the video presenting the verb "to rain" (pol. *padać*). The frame marked with the black box, which takes place in the moment when the hand is fully raised just before it comes back down, could be considered the one that presents the initial configuration. Unfortunately, this approach cannot be applied to every sign. In Fig. 3(b), where the frames of the sequence presenting the noun "future" (pol. *przyszłość*) are gathered, the trajectory of the hand does not change in the $y$ axis, hence the rule discussed earlier cannot be applied in this case. Moreover, since each sign starts from the neutral position (see Fig. 3(a) – first frame) or immediately after the previous sign ends, the lack of the unified definition of the initial sign configuration complicates the implementation of automatic detection of the initial position of a signer.

(a) to rain (pol. *padać*)

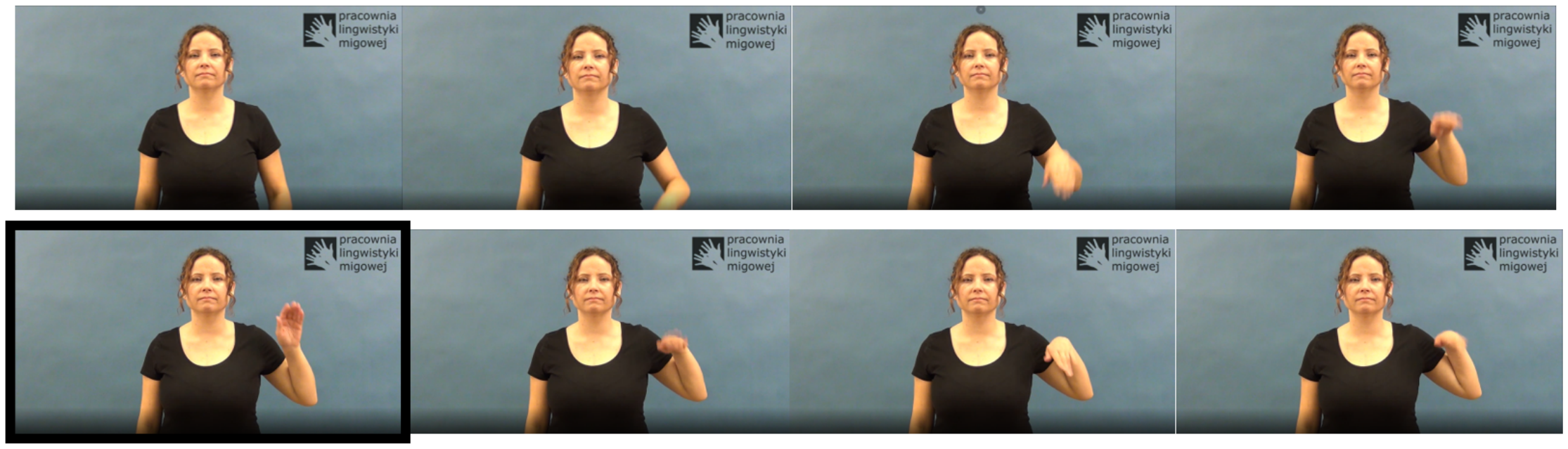

(b) future (pol. *przyszłość*)

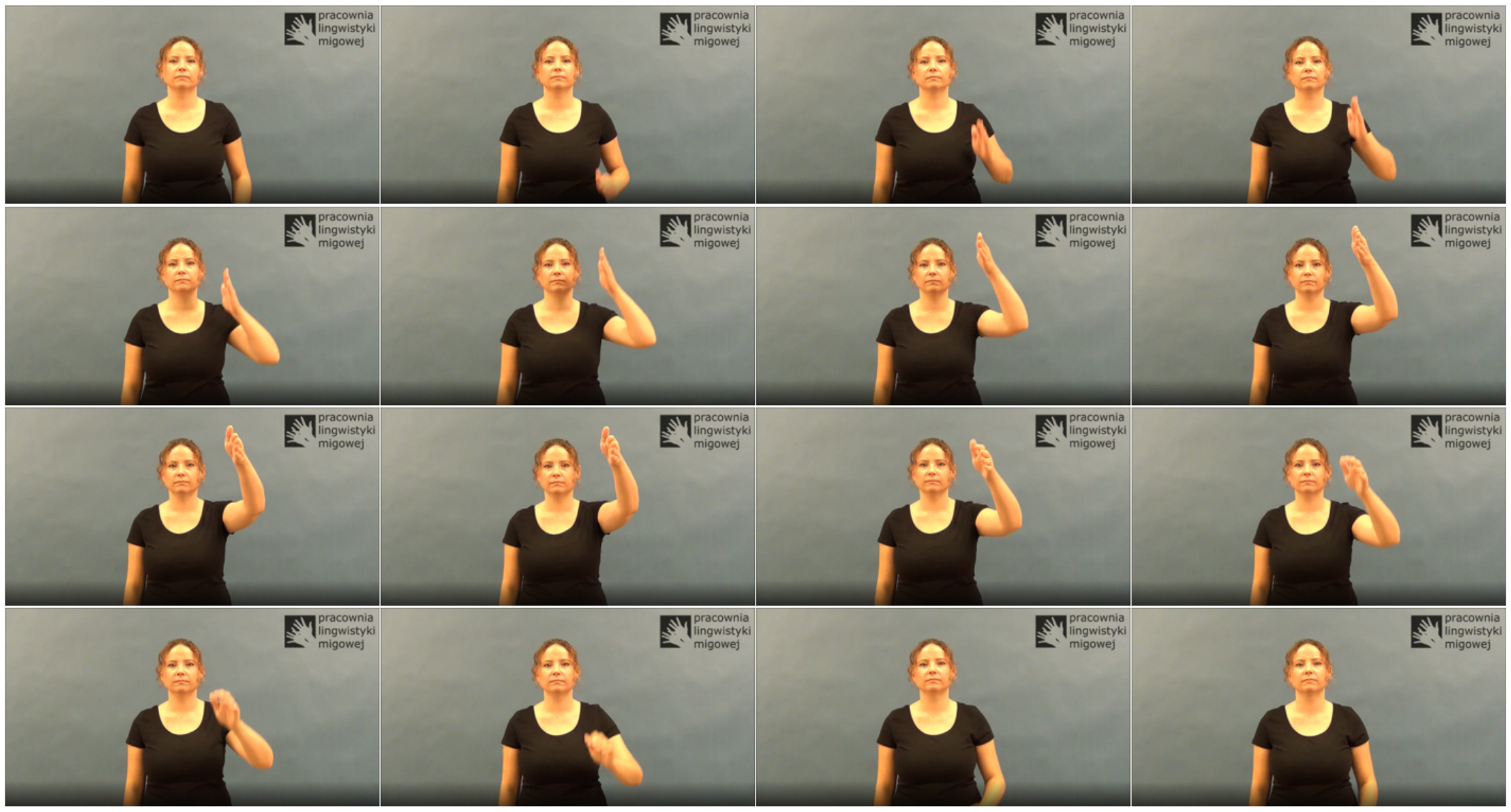

Figure 3: Visualization of the challenge to select a frame presenting the signer's initial configuration for two words from the Corpus-based Dictionary of Polish Sign Language [15]: (a) the verb "to rain" (pol. *padać*) (b) the noun "future" (pol. *przyszłość*). For case (a), the inflection point is considered the initial position (bold black frame), in case (b) the same methodology cannot be successfully applied.

Given that the datasets analyzed in this work were created by different groups, we cannot assume that the same procedure was used to annotate them. Therefore, selecting the initial frame was fully up to the annotator and lacks a standardized procedure. Nevertheless, in order to automatically analyze the videos of signs in sign language, we needed to automatically identify the frame with the initial position of the signer. To determine it, we utilized the MediaPipe Pose coordinates (landmarks), as described in Section 3.4. We then identified the first local extrema in the $x$- and $y$-axis trajectories for three landmarks: the wrist, thumb carpometacarpal (CMC), and thumb metacarpophalangeal (MCP) joints, obtaining twelve values of the potential initial position frame for each hand (see Fig 4). The frames range might differ for each hand as landmarks can be visible in the video for a different period. Taking into account the possibility of inaccuracy in pose estimation, we took their median value. We used only the landmarks located on the hands of the signer, as they are the most distinctive and represent the most variable movement.

The analysis of the results allowed us to conclude that the selected number of points is sufficient to determine the initial position frame. However, it is worth mentioning that in some cases the inflection point of the hand trajectory was not detected due to the nature of the movement, e.g., the sign is made in a different plane or the selected landmark is not visible during the movement.

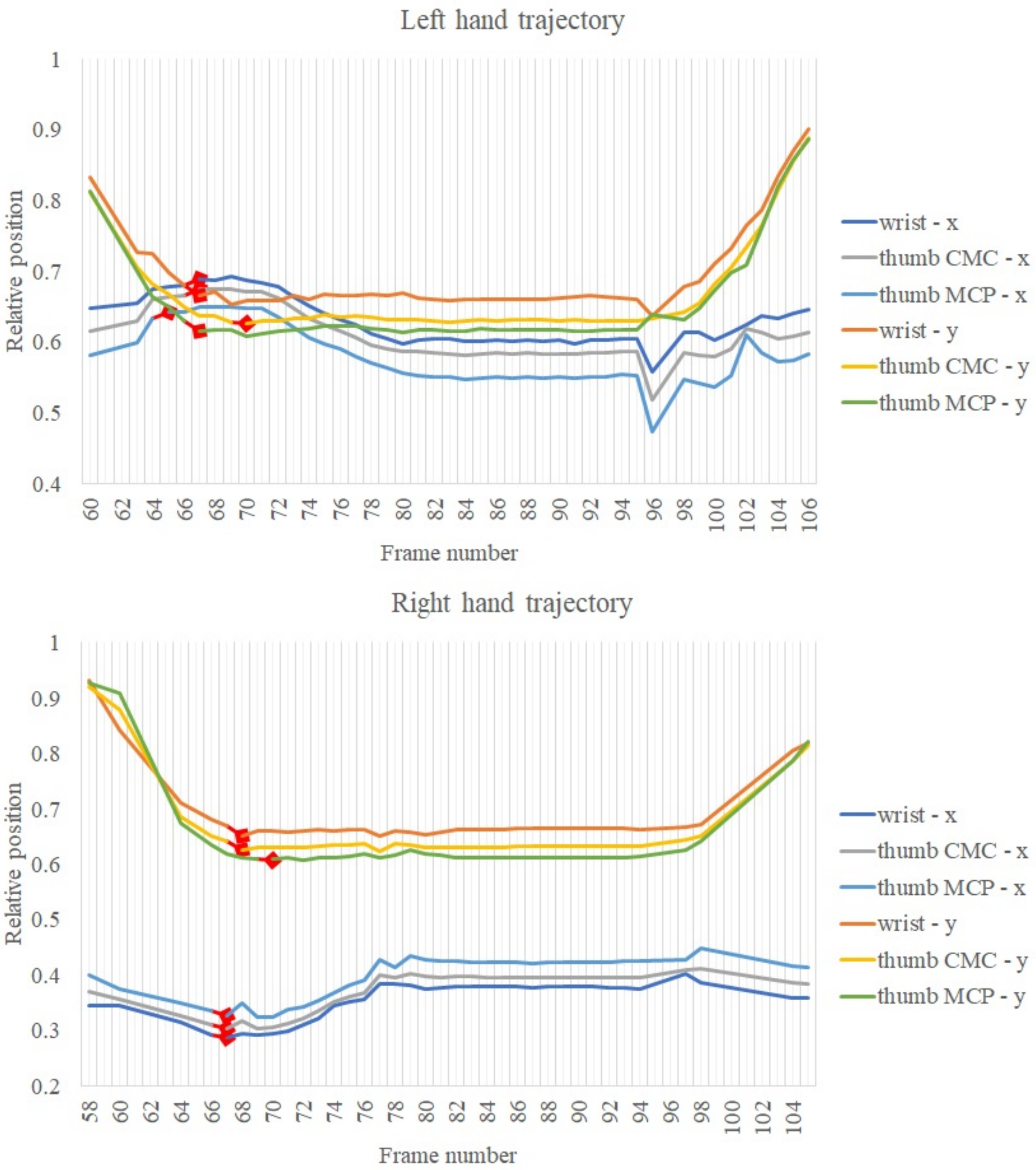

Figure 4: The visualization of the initial position frame determination process. The initial position frame is indicated by the first local extremum, which is shown as a red marker, and by the median value for each of the landmarks (wrist, thumb CMC, and thumb MCP) taken into consideration.

### 4.2 Distinction between dominant and non-dominant hand

Sign languages use the visual channel for communication. Multiple articulators can be involved in the generation of all phonological material, including one or both hands, as some signs require one hand (one-handed signs), while others have a specification for two hands (two-handed signs). Since the choice of hands between "left" and "right" has not been shown to play a linguistic role in any sign language, the distinction between two hands is described in terms of the "dominant" vs. the "non-dominant" hand.

The HamNoSys annotation guidelines do not clearly define the rules for the distinction between those two hands. This must be deduced in some way, but it is challenging to develop a set of guidelines that will work for all real-life cases. Onno Crasborn [50] defined the dominant hand as the hand that is most active with respect to the other one. His definition states that the dominant hand is normally above the non-dominant in signs where both hands move and one is higher than the other.

Another factor that complicates hand determination is a dominance reversal [50] – change of dominant hand, which may occur when from a practical point of view, another hand, usually a non-dominant one, is more convenient to use. It is important to mention that signers can also switch dominance according to their convenience or some linguistic reasons.

The lack of clear and unified rules combined with a possible dominance reversal and limited data access completely precludes correct evaluation of a hand type that should refer to the dominant hand. Also, the symmetry of the movement is defined regarding this hand. Consequently, this can lead to decisions based only on the annotator's interpretation. Hence, annotations might be inconsistent between different corpora or even within a single corpus.

### 4.3 Ambiguity in determining the hand location

Many HamNoSys transcriptions appear to lack consistency in terms of accurate notation of hand location. This can be a problem of HamNoSys itself, which tries to provide a very detailed description of the vertical position – $y$- axis position – to which we refer when determining the $x$- and $z$-axis positions. Also, the lack of a strict rule defining the signer's initial position affects the quality level of the provided labels. Another aspect that raises doubts about the notation homogeneity of hand location is related to the number of its possible classes. According to Table 2, we encoded 36 possible classes of *hand location top/bottom*, of which as many as 15 specify only the face area. Although this kind of description is very specific, it is hard to reproduce in the exact same way and not confuse it with other classes in practice.

We used Mediapipe Pose to automatically determine the position of the hand based on the selected keypoints. Next, the geometric relationships between several landmarks can be utilized to examine gesture variability for various ground-truth annotations. For instance, if the hand location on the selected initial frame is the same for several symbols of the HamNoSys, vision-based computer systems may find it difficult to distinguish between them.

The measurement results for evaluating the labeling consistency for various hand location classes are presented in this section. Based on the measured distances in pixels between the nose keypoint and the wrists of the right/left hand, we determined how much the predicted length differs within the same class. We focused on a few classes of the *location top/bottom* in the central position for the signer's initial configuration frame that were chosen using the method described in Section 4.1. The derived average distances, given as the $L_1$ and $L_2$ metrics are summarized in Table 3, along with the standard deviation for each class. Since the videos inside one corpus have a similar acquisition methodology and resolution, the calculations are carried out separately for each dataset, thereby ensuring the precision of the calculations reported in pixels. The missing values in the table represent the absence of specific classes in the dataset under analysis.

According to the findings for the right hand in Table 3, the standard deviation within a class is a value within the range of 1 (head top) and 71 (chin) percent of the mean measured distance. Not surprisingly, the average distance for the measured distance is very similar for closely located assigned body parts. There is just one instance of a specific class in the dataset and there is no estimated standard deviation for that class. This may suggest that some classes are redundant and difficult to annotate appropriately, such as distinguishing the head and eye symbol (see Figure 5). Additionally, there may be some inconsistency due to the disparity between the dominant and non-dominant hand. This is particularly visible for the CDPSL corpus, where standard deviations are the largest – it was observed both for right and left hand. Such specifics may result from the tendency of different sign languages to dynamically shift hands during the process of presenting a sign.

However, we must remember that for the $x$-axis, the distance of the right hand from the nose can vary as well. The aforementioned analyses were repeated for the *left/right-hand location* block (6 classes) using the neutral vertical position of the hand at the breast level to test the precision of the annotation for the horizontal position. The achieved results are summarized in Table 4. All the creators of the corpora under study only occasionally employ the position that suggests moving the hand far to the left in the breast line in the initial configuration. The most significant standard deviations were found in this scenario for the basic lexicon, which is a combination of multiple sign languages. The overall outcome clearly shows a notable inconsistency with the category's class identification.

Table 3: Statistics for the analyzed distance between the right wrist and nose for selected classes of *hand location top/bottom*. We report an average ± standard deviation. The statistics are calculated in pixels - separate for each dataset.

| Class | Norm | GLEX | GALEX | BL | CDPSL | GSLL |
|---|---|---|---|---|---|---|
| | $L_1$ | 9.41 ± 3.20 | 22.23 ± 2.07 | 133.13 ± 44.67 | 334.22 ± 208.07 | 350.18±26.93 |
| head top | $L_2$ | 6.90 ± 2.24 | 17.81 ± 0.17 | 112.67 ± 42.75 | 277.95 ± 160.77 | 264.48 ± 32.75 |
| | $L_1$ | 19.32 ± 6.35 | 41.11 ± 19.69 | 123.83 ± 41.43 | 279.61 ± 150.56 | 397.41 ± 104.33 |
| head | $L_2$ | 15.01 ± 4.57 | 32.38 ± 14.41 | 103.23 ± 38.17 | 219.10 ± 122.19 | 322.18 ± 106.57 |
| | $L_1$ | — | — | — | 236.51 ± 151.35 | 323.79 ± 161.49 |
| eyes | $L_2$ | — | — | — | 177.82 ± 121.52 | 263.14 ± 142.70 |
| | $L_1$ | 30.46 ± 10.18 | 29.49 ± 14.34 | 29.49 ± 14.34 | 386.15 ± 259.02 | 384.21 ± 150.65 |
| nose | $L_2$ | 23.07 ± 7.93 | 22.15 ± 12.25 | 22.15 ± 12.25 | 300.28 ± 209.88 | 335.89 ± 153.70 |
| | $L_1$ | 34.27 ± 8.72 | 30.23 ± 5.98 | 30.23 ± 5.98 | 413.78 ± 236.85 | 370.75 ± 135.20 |
| chin | $L_2$ | 26.95 ± 7.21 | 23.54 ± 4.58 | 23.54 ± 4.58 | 328.05 ± 191.11 | 320.20 ± 128.22 |
| | $L_1$ | 30.86 ± 11.60 | 30.42 | 124.56 ± 37.01 | 355.54 ± 245.65 | 434.87 ± 124.10 |
| under chin | $L_2$ | 24.33 ± 9.53 | 22.97 | 106.91 ± 32.78 | 279.95 ± 197.58 | 383.21 ± 115.85 |
| | $L_1$ | 48.70 ± 26.26 | 36.75 ± 10.85 | 131.87 ± 35.48 | 313.66 ± 162.76 | 440.70 ± 104.69 |
| neck | $L_2$ | 39.90 ± 20.95 | 32.51 ± 9.23 | 113.30 ± 32.12 | 256.90 ± 127.00 | 386.12 ± 100.75 |
| | $L_1$ | — | 47.39 | 145.67 ± 30.33 | 452.28 ± 207.74 | 412.46 ± 114.14 |
| top of shoulder | $L_2$ | — | 34.29 | 120.90 ± 28.02 | 354.37 ± 173.94 | 348.02 ± 110.48 |
| | $L_1$ | 38.51 ± 8.12 | 36.78 ± 9.43 | 138.21 ± 27.65 | 393.68 ± 150.27 | 420.91 ± 100.99 |
| shoulder line | $L_2$ | 29.05 ± 6.49 | 28.22 ± 6.88 | 114.19 ± 27.23 | 312.18 ± 122.94 | 356.95 ± 97.41 |
| | $L_1$ | 50.32 ± 10.20 | 50.78 ± 8.81 | 130.81 ± 29.95 | 469.10 ± 162.44 | 453.33 ± 93.21 |
| breast line | $L_2$ | 39.77 ± 7.96 | 39.17 ± 6.51 | 107.30 ± 27.86 | 365.99 ± 133.30 | 384.44 ± 88.59 |
| | $L_1$ | 61.00 ± 7.89 | 60.57 ± 11.64 | 149.57 ± 27.47 | 486.54 ± 154.73 | 482.21 ± 82.56 |
| belly line | $L_2$ | 49.78 ± 5.74 | 49.83 ± 1.12 | 123.73 ± 23.39 | 379.48 ± 125.48 | 409.03 ± 84.83 |
| | $L_1$ | 69.69 ± 6.41 | — | 129.54 ± 13.89 | 414.37 ± 218.10 | — |
| abdominal line | $L_2$ | 57.57 ± 3.14 | — | 107.57 ± 22.86 | 348.22 ± 185.28 | — |

Table 4: Statistics for the analyzed distance between the right wrist and nose for different classes of *hand location left/right*. The HamNoSys characters are provided in reference to top/bottom location – breast line. We report average ± standard deviation. The statistics are calculated in pixels - separate for each dataset.

| Class | Norm | GLEX | GALEX | BL | CDPSL | GSLL |
|---|---|---|---|---|---|---|
| | $L_1$ | 50.32 ± 10.20 | 50.78 ± 8.81 | 130.81 ± 29.95 | 469.10 ± 162.44 | 453.33 ± 93.21 |
| center of | $L_2$ | 39.77 ± 7.96 | 39.17 ± 6.51 | 107.30 ± 27.86 | 365.99 ± 133.30 | 384.44 ± 88.59 |
| | $L_1$ | — | — | 145.20 | 510.65 | — |
| left to | $L_2$ | — | — | 124.93 | 361.20 | — |
| | $L_1$ | 42.97 ± 13.40 | 36.40 | 129.02 ± 30.49 | 376.64 ± 176.43 | 412.03 ± 79.19 |
| left side of | $L_2$ | 37.41 ± 7.58 | 35.26 | 107.39 ± 27.74 | 311.50 ± 129.03 | 391.42 ± 71.41 |
| | $L_1$ | 68.92 ± 7.25 | 61.98 ± 13.28 | 154.94 ± 30.18 | 538.67 ± 158.9 | 424.87 ± 19.56 |
| right to | $L_2$ | 51.08 ± 4.67 | 46.27 ± 7.30 | 126.78 ± 24.92 | 400.80 ± 118.09 | 384.96 ± 32.034 |
| | $L_1$ | 54.29 ± 11.63 | 52.46 ± 9.50 | 148.16 ± 28.61 | 490.07 ± 142.83 | 466.51 ± 89.95 |
| right side of | $L_2$ | 41.75 ± 7.09 | 39.96 ± 6.57 | 121.07 ± 26.50 | 374.95 ± 118.51 | 388.78 ± 87.48 |

Another feature of hand location that is problematic since it is not well defined is the *distance*. As presented in Table 2 hand location distance class, that shall be considered in the initial position defines 7 possible distances. It describes the relationship between the dominant hand and the other part of the body that takes part in the sign. Unfortunately, a class named *Standard distance* is more intuitive than strictly defined, meaning that the choice of the distance definition is, again, entirely up to the annotator. It is essential to mention, that if the distance is omitted in the annotation, standard/natural distance is assumed. To visualize the problem, we presented a few samples of initial position configurations that are confusing in Figure 6. For instance, hands in the words 'Australia' and read' have similar distances from each other, but in the first case, the distance is omitted and assumed as "standard", while in the second defined as close (⁾⁽). Although it might be connected to the head (○) character, which appears after the distance character, it is not clear what exactly the distance refers to. In turn, the word 'long-time' shown in Figure 6(c) has a hand distance close to body (⁾⁽) after the *location top/bottom* upper arm (ʅ). However, it is hard to distinguish from standard distance from the body. It is worth noticing that in Figure 6(b) the distance is placed before the body part, while in Figure 6(c) after it. Finally, in Figure 6(d) the word 'good' is presented. Although there is no distance specification, which indicates the standard distance, the signer visibly touches her cheek with her fingers.

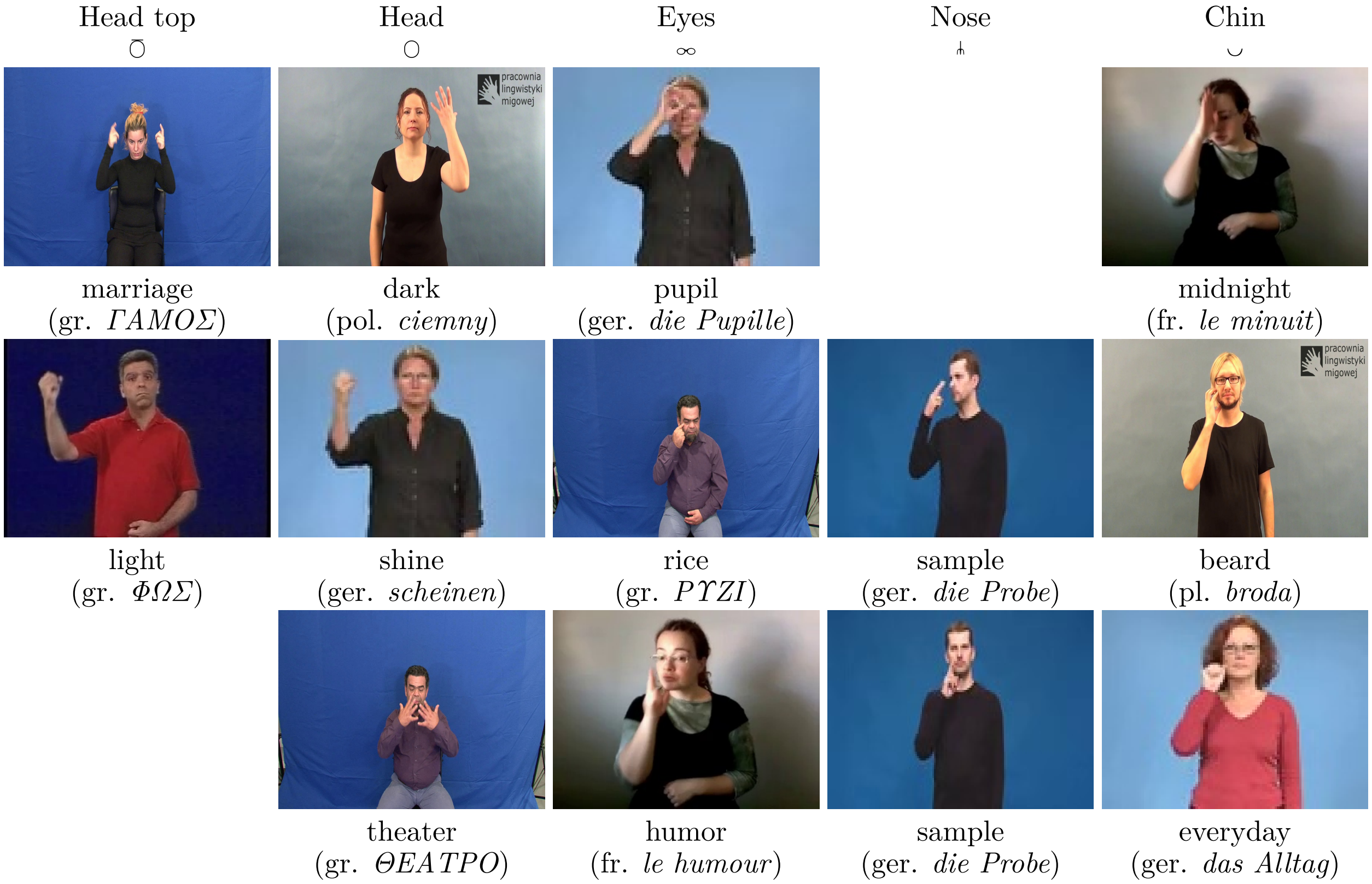

Figure 5: Inconsistency in the initial configuration hand location top/bottom for analyzed corpora. Although the hand is positioned at similar height in each row (for different corpora), the provided labels belong to a distinct classes.

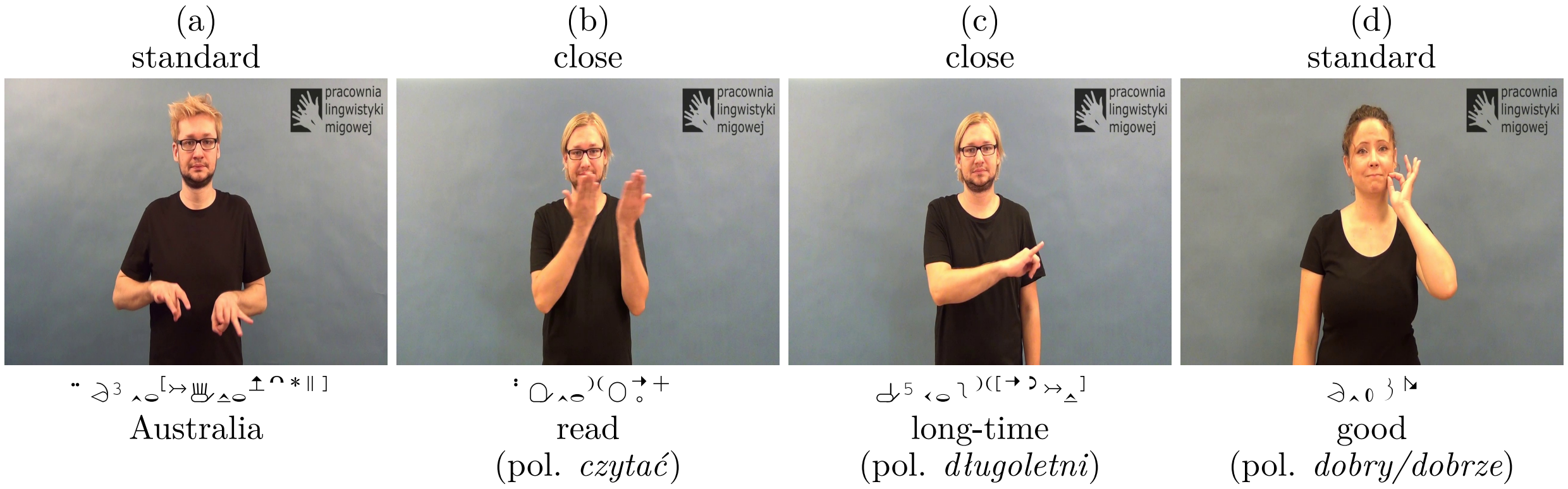

Figure 6: Examples of inconsistency in the initial configuration *hand location distance* for selected glosses from CDPSL [15].

### 4.4 Subject bias

Subject bias is a behavior of experiment participants that involves acting the way that they think, they are supposed to act, and mostly appears when subjects know the purpose of the study [56]. In case of signs annotation, using HamNoSys, the label might be influenced by the annotator's knowledge. There might be a difference between what is visible in the video and what the sign should look like by the book.

Sign languages are closely related to the place of use – they do not only depend on the country but even on the inland region. The same word can be signed in various ways, even within the same sign language. Additionally, groups of people may use their own colloquial language, therefore there might be some differences between the knowledge of the annotator and the signer. This can lead to some inconsistencies in sign language video annotations.

Figure 7 shows three different ways to relate to the word "see" as it is used in DGS [14] and PJM [15], along with their HamNoSys labels. In the case of DGS gloss *sehen* (see Figure 7(a)), the dominant hand is placed at the signer's nose level, which corresponds to the proper HamNoSys symbol . The same position has been marked as and in both cases of the identical word shown in PJM (see Figure 7(b-c)), despite the hand being at a similar height. This variance may be caused by various regional or national contexts, as well as the semantic association of the word with the eyes – the sense organ of sight.

(a) nose | (b) under nose | (c) eyes

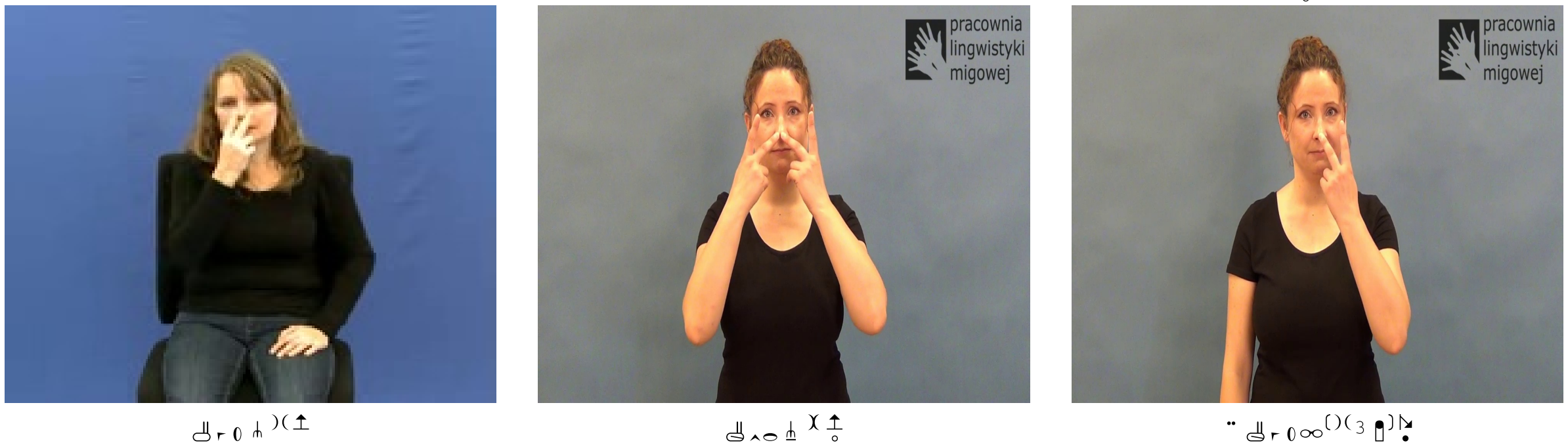

Figure 7: Different initial positions for word "see" in (a) DGS [14] and (b-c) PJM [15].

### 4.5 Impact of the reduction of HamNoSys symbols

In the final stage of our analysis, we encoded all the gathered data (described in Section 3.2) using our parser and examined its key statistics. The number of examples we have for each HamNoSys class is shown in Figure 8. The order in which the HamNoSys symbols are presented in Table 2 determines how the symbols were encoded into numbers. Firstly, it is important to note the significant variation in class distribution across each block. In addition, the number of instances per class inside each block is very imbalanced. There were no cases found for several categories, such as the class 10 () of the *Extended finger direction* block, which corresponds to the index direction of the hand pointing to the bottom right side from a bird's eye view. This may be because not every hand position is natural, which resulted in fewer instances of the specific symbols (particularly when combined with others).

It is important to keep in mind that the use of the parser significantly decreased the number of symbols in the evaluated labels. We believe that a given reduction did not significantly impair the understanding of a given gesture from the perspective of data preparation for the machine learning model training. To test this hypothesis we performed backward decoding from the designated numeric multilabels to the appropriate glosses. The study aimed to investigate whether it is possible to recreate a sign using a reduced parsed HamNoSys label. If the parser assigned one HamNoSys label to each tested gloss, we interpreted it to indicate that the decoding procedure was successful. The verification is deemed ineffective if it assigned more than one label. Due to the diversity and varying amounts of data in each database, the results were independently verified for each dataset.

The impact of data reduction on each dataset is summarized in Table 5. The Basic Lexicon dataset consisted of four sign languages and has the lowest decoding efficiency $\eta = \frac{\#singly\,labeled}{\#unique\,glosses}$. The maximum number of glosses that were mistakenly associated with a single HamNoSys label in this database is 12. The event assumed to be a success (one HamNoSys label is one gloss) was not always accurate, especially when numerous sign languages were evaluated at once [11]. The parser was able to accurately detect a single gloss from distinct SLs even for reduced numerical multilabels, as evidenced by the predicted high decoding efficiency.

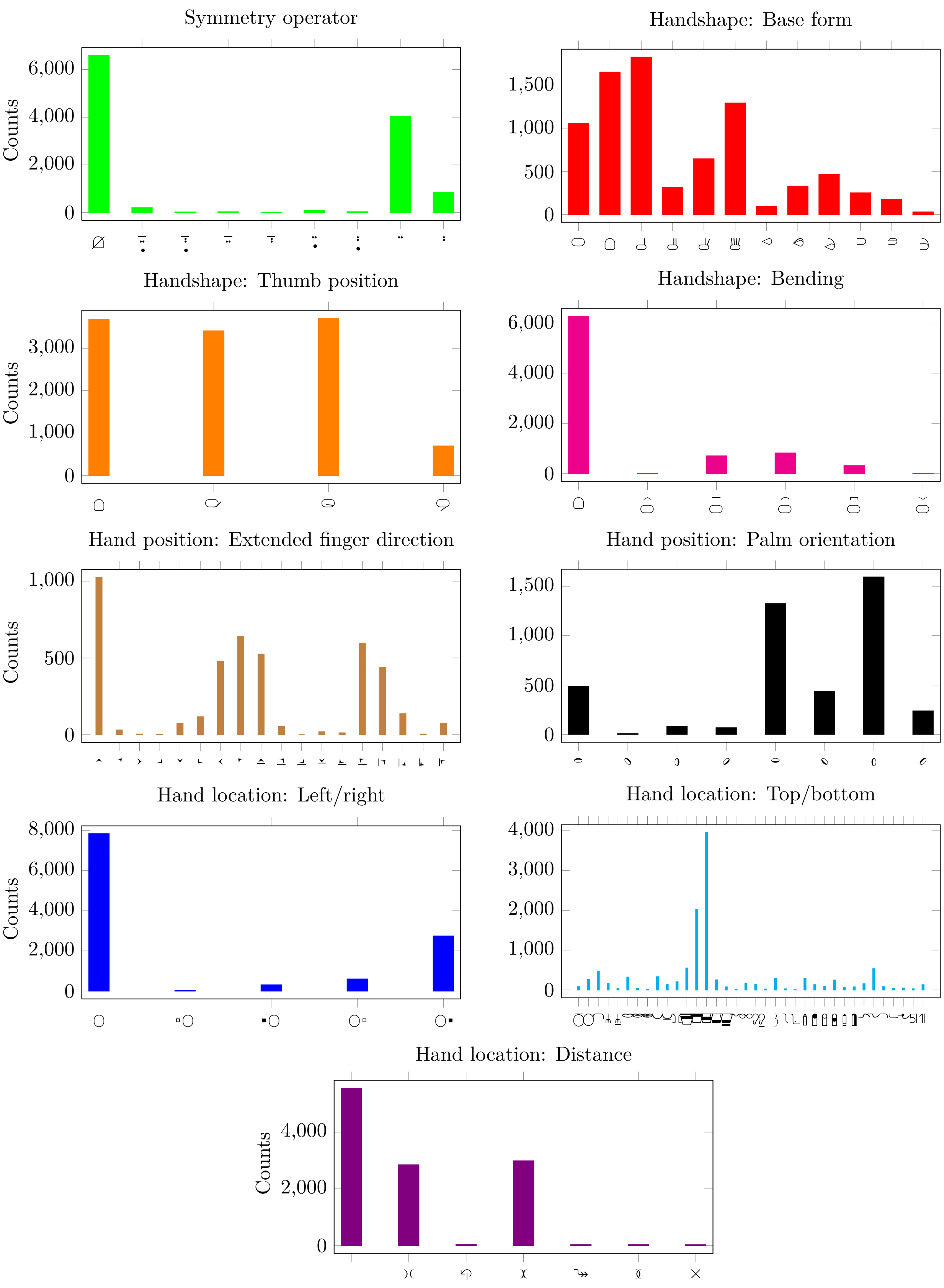

Figure 8: Statistics for the data used in the isolated sign language recognition task – the number of glosses for each analyzed HamNoSys symbol.

In light of the aforementioned investigation, we drew the conclusion that data reduction has little effect on this procedure and may rather benefit automatic translation systems [54].

Table 5: A summary of the impact of data reduction on individual datasets. Decoding efficiency $\eta$ shows the parser's capability to recognize a single isolated gloss.

| **Dataset** | **GLEX** | **GALEX** | **CDPSL** | **BL** | **GSLL** |
|---|---|---|---|---|---|
| $\eta$ (%) | 94.61 | 94.16 | 91.09 | 83.82 | 94.33 |

Another factor, important from the machine learning point of view is an unbalanced number of instances per class, as can be seen in Fig. 8. For instance, such ten classes of the *Extended finger direction* group, which corresponds to the index direction of the hand pointing to the bottom right side from a bird's eye view, no case was observed. Such data distribution may strongly affect the machine learning model training. The problem of class imbalance is pretty common. Although there are some methods for overcoming this issue, not enough representation of classes would still be a significant problem.

## 5 Discussion

Supervised machine learning approaches fully rely on labeled data to train models, and their capabilities are directly affected by the quality of such ground-truth data. One of the main concerns is the quality, uncertainty, and subjectivity of the provided labels. It is well known that even for seemingly simple tasks such as sentiment analysis, human annotators sometimes disagree, which makes the labels of the data inconsistent [57]. This, in turn, results in poor model performance and generalization abilities over unseen data. The question remains – how can we make sure we can trust the labels in a dataset and thus remove the issue of label noise? One of the solutions is a careful analysis of the used datasets in terms of potential label inconsistencies, noise, missing or vague information, and other related data-level issues. In this paper, we approached this important problem and performed the HamNoSys analysis in terms of usability for automatic sign language processing and translation.

The major issue which was identified is that HamNoSys was created in order to unify the notation method of various sign languages, not for computer-aided systems synthesis. As long as some hidden, omitted or default information in HamNoSys notation might be obvious for a human reader, decoding it by a computer might be a challenge.

In addition, several label components are defined relative to other symbols, thus if one assignment was incorrect, the entire gesture would be encoded incorrectly. Stricter and more accurate regulations for sign annotations bundled in a standardized annotation procedure would be the only way to effectively address this issue.

A similar problem may apply to the selection of the signer's initial position, which is not defined by any rule and there is an implicit assumption that the initial position is known to the annotator based on the knowledge of the analyzed sign language. The proposed initial position frame selection might be a solution to this problem. Supporting annotators by a computer-aided system could also appear beneficial for further training, as the labels would be coherent with the chosen fragment of a video.

There is also a subject bias that strongly influences annotations. For example, the HamNoSys notation provides information about the dominant hand for a sign, but in practice, the selection of the dominant hand is fully up to the signer. As a result, this information is hard to forward in a unified data labeling protocol.

Nevertheless, the influence of habits may play a significant role in the annotation process. This also leads to another consideration – how exactly the data should be annotated: from the perspective of video recordings or common knowledge. We proved in the paper that sign languages, similarly to verbal ones are not mechanical and strongly depend on the signer. Therefore, video recordings might be influenced by some signer-dependent factors. In contrast, labeling based on the dictionary knowledge may lead to some inconsistencies between expected HamNoSys character and signer position or movement and thus poor performance of the data-driven models. However, we can compare it to some inconsistencies in voice recordings for natural language processing, such as speech defects. Therefore, we assume that labels should be created according to the dictionary knowledge and all the inconsistencies treated as noise.

Another concern regarding HamNoSys-based labels is the large number of classes in each block. For instance, in the position of the hand in the vertical axis (the *location top/bottom*) there are above 35 possible variants.

When we consider two close classes, for example, the *upper lips* and the *upper teeth row*, it is impossible for different annotators to consistently annotate a given sign without dictionary knowledge. At the same time, the positioning assumes that the hand is a single point, in which case it is not entirely clear whether the symbol on the label refers to the wrist or the fingertips. Furthermore, it significantly hinders the training of a machine learning model, as there would be no visible difference between those classes. To reduce this problem a huge amount of data would be necessary.

Although Richard Kennaway [23] discussed that HamNoSys omits some necessary details that may affect the correctness of the artificially generated avatar's representation of the gesture and thus modify its meaning, our analysis involving backward decoding of the numerical labels extracted from HamNoSys to gloss showed that the losses obtained in this process can be negligible. Further, we also calculated the distance between the chosen landmarks for the signer's initial configuration. It turned out that the standard deviation values in each sign are so high, that labels are poor and do not allow for diversification between similar signs.

It is also worth analyzing HamNoSys-based labels as a whole string, not separate sub-units. We could get inspired by Transformers used for natural language processing as contextual information is crucial for translation [6]. Similarly in HamNoSys, the characters relate to each other, so maybe their analysis should be global. This, however, would also require a large database.

It can be noted, that some of the mentioned problems could be reduced by the usage of a greater database. Most of the data used in this research were taken from dictionaries, which means there is only one representation for each gloss and only one or few signers are involved. Next, there are multiple ways for signing some glosses. They probably slightly differ in meaning, however, in the dictionary, it introduces confusion. It is commonly known that such a database creation is difficult and expensive, so the labeling process should be as simple and consistent as possible.

## 6 Conclusions

In this work, we investigated an array of potential issues concerned with the phonetic transcription labels in sign language resources. The objective of our research was to draw attention to potential errors and biases in the usage of the language-neutral Hamburg Sign Language Notation System (HamNoSys) for machine learning labeling purposes, as well as to the other problems that might negatively impact the performance of data-driven machine learning algorithms benefiting from such ground-truth information during their training.

Despite the fact that HamNoSys is very useful for standardizing the transcription of various Sign Languages, it has significant shortcomings caused by its several underlying inconsistencies. Here, we mainly focused on non-linguistic problems that can obstruct computer-aided support for annotation, recognition, and translation of any sign language. Due to a lack of standardized annotation procedures, the difficulties in the labeling process based on HamNoSys notation lead to inconsistencies and therefore cannot be used to train machine learning models.

The work reported in this paper constitutes an interesting point of departure for further research. None of the techniques discussed in the 2 provided mitigation strategies for enabling automated labeling using HamNoSys. Both manual and automatic labeling involves some noise and bias in the provided labels. Therefore, future work around phonetic translation systems might focus on the creation of unified and standardized guidelines for data annotation. In addition, access to a huge, diversified database would lead to better generalization ability and would at least reduce the identified problems.

## Acknowledgments

The five-month non-profit educational project HearAI[4] was organized by Natalia Czerep, Sylwia Majchrowska, Agnieszka Mikołajczyk-Bareła, Milena Olech, Marta Plantykow, Żaneta Lucka Tomczyk. The main goals were to work on sign language recognition using HamNoSys, and to increase public awareness of the Deaf community. The authors acknowledge team members: Natalia Czerep, Maria Ferlin, Wiktor Filipiuk, Grzegorz Goryl, Agnieszka Kamińska, Alicja Krzemińska, Adrian Lachowicz, Sylwia Majchrowska, Agnieszka Mikołajczyk-Bareła, Milena Olech, Marta Plantykow, Patryk Radzki, Krzysztof Bork-Ceszlak, Alicja Kwasniewska, Karol Majek, Jakub Nalepa, Marek Sowa, Wiktoria Choręziak, Jacek Kawalec, Maria Lewalska, Anna Monik, Żaneta Lucka Tomczyk, Adela Rużyła-Wrzeszcz who contributed to the project.

[4]https://www.hearai.pl/

The authors acknowledge the infrastructure and support of Voicelab.AI in Gdańsk, and financial support of Polish Development Fund (PFR) Foundation.